\documentclass[twosides]{article}

 \usepackage{RR}

\usepackage{times}
\usepackage{helvet}
\usepackage{courier}
\usepackage[utf8]{inputenc} % allow utf-8 input
\usepackage[T1]{fontenc}    % use 8-bit T1 fonts
\usepackage[usenames,dvipsnames]{xcolor}
\definecolor{pdfurlcolor}{rgb}{0,0,0.6}
\definecolor{pdfcitecolor}{rgb}{0,0.6,0}
\definecolor{pdflinkcolor}{rgb}{0.6,0,0}
\usepackage[colorlinks=true,citecolor=pdfcitecolor,urlcolor=pdfurlcolor,linkcolor=pdflinkcolor,pdfborder={0 0 0}]{hyperref}

\usepackage{graphicx}

\usepackage{booktabs}       % professional-quality tables
\usepackage{amsfonts}       % blackboard math symbols
\usepackage{nicefrac}       % compact symbols for 1/2, etc.
\usepackage{microtype}      % microtypography
\usepackage{esdiff}
\usepackage{multirow}

\usepackage{stmaryrd}
\usepackage{enumerate}
\usepackage[cmex10]{amsmath}
\usepackage{bm}
\usepackage{amsfonts,amssymb,amsthm,amstext,latexsym}
\usepackage{xspace}
\usepackage{paralist}
\usepackage[normalem]{ulem}
\usepackage[nottoc]{tocbibind}
\usepackage{amsmath}

\usepackage{subcaption}

\usepackage{algorithm}
\usepackage{algpseudocode}

\usepackage{tikz}
\usepackage{todonotes}

\usepackage{graphicx}
\graphicspath{{./plots/}}

\usetikzlibrary{patterns}
\usetikzlibrary{calc}
\usetikzlibrary{decorations.pathmorphing} % noisy shapes
\usetikzlibrary{fit}					% fitting shapes to coordinates
\usetikzlibrary{backgrounds}	% drawing the background after the foreground

\usetikzlibrary{arrows,shapes}
\usetikzlibrary{trees,snakes}
\usetikzlibrary{decorations.markings}
\usetikzlibrary{shadows}

\theoremstyle{plain}

\newtheorem{theorem}{Theorem}

\theoremstyle{definition}

\newcommand{\ema}[1]{\ensuremath{#1}\xspace}

\newcommand{\optBP}[1][{s, t, m}]{\ema{C_{\text{BP}}\left(#1\right)}}
\newcommand{\uf}[1][]{\ema{u_f^{#1}}}
\newcommand{\ub}[1][]{\ema{u_b^{#1}}}
\newcommand{\of}[1][]{\ema{o_f^{#1}}}
\newcommand{\ob}[1][]{\ema{o_b^{#1}}}

\newcommand{\wx}[1][]{\ema{\omega_a^{#1}}}
\newcommand{\wy}[1][]{\ema{\omega_\delta^{#1}}}
\newcommand{\wbx}[1][]{\ema{\omega_{\bar{a}}^{#1}}}
\newcommand{\nomem}[1][]{\ema{m_{\varnothing}^{#1}}}
\newcommand{\emem}[1][]{\ema{m_{all}^{#1}}}

\newcommand{\forop}[1][]{\ema{F_{#1}}}
\newcommand{\eforop}[1][]{\ema{F_{all}^{#1}}}
\newcommand{\nforop}[1][]{\ema{F_{\varnothing}^{#1}}}
\newcommand{\cforop}[1][]{\ema{F_{ck}^{#1}}}

\RRtitle{Checkpointing optimal pour chaînes hétérogènes: apprentissage de réseaux de neurones profonds avec mémoire limitée}

\RRetitle{Optimal checkpointing for heterogeneous chains: how to train deep neural networks with limited memory}
\titlehead{Optimal checkpointing for heterogeneous chains}

\RRauthor{Olivier Beaumont\thanks[a]{Inria Bordeaux -- Sud-Ouest and Universit\'e de Bordeaux} \and Lionel Eyraud-Dubois\thanksref{a} \and Julien Hermann\thanksref{a} \and Alexis Joly\thanks{Inria Sophia-Antipolis M\'editerran\'ee and Universit\'e de Montpellier} \and Alena Shilova\thanksref{a}}

\RRprojets{HiePACS and Zenith}
\RCBordeaux
\RRkeyword{Deep Learning, Machine Learning, Scheduling, Checkpointing,
  Automatic Differentiation}
\RRmotcle{Apprentissage Profond, Réseaux de Neurones, Ordonnancement,
  Checkpointing, Différentiation Automatique}
\RRdate{November 2019}
\RRNo{9302}

\RRabstract{
This paper introduces a new activation checkpointing method which
allows to significantly decrease memory usage when training Deep
Neural Networks with the back-propagation algorithm.  Similarly to
checkpointing techniques coming from the literature on Automatic
Differentiation, it consists in dynamically selecting the forward
activations that are saved during the training phase, and then
automatically recomputing missing activations from those previously
recorded.  We propose an original computation model that combines two
types of activation savings: either only storing the layer inputs, or
recording the complete history of operations that produced the outputs
(this uses more memory, but requires fewer recomputations in the
backward phase), and we provide an algorithm to compute the optimal
computation sequence for this model, when restricted to memory
persistent sequences.

This paper also describes a PyTorch implementation that 
processes the entire chain, dealing with any sequential 
DNN whose internal layers may be arbitrarily complex and
automatically executing it according to the 
optimal checkpointing strategy computed given a memory limit. 
Through extensive experiments, we show that our implementation consistently 
outperforms existing checkpointing approaches for a large class of networks, image sizes and batch sizes.
}

\RRresume{ Cet article introduit une nouvelle méthode de sauvegarde
  des activations qui permet de réduire significavement la mémoire
  utilisée lors de la phase d'apprentissage de Réseaux de Neurones
  Profonds avec l'algorithme de rétropropagation. Cette méthode,
  inspirée des techniques de checkpoint en Différentiation
  Automatique, sélectionne dynamiquement les activations sauvegardées
  pendant la phase avant, puis recalcule automatiquement les
  activations manquantes à partir de celles sauvegardées
  précédemment. Nous proposons un modèle de calcul original qui
  combine deux façons de sauvegarder une activation : soit ne stocker
  que les entrées de la couche concernée, soit enregistrer
  l'historique complet des opérations qui ont permis de produire les
  sorties (cela utilise plus de mémoire, mais nécessite moins de
  recalcul dans la phase arrière). Nous présentons un algorithme qui
  fournit la séquence de calculer la séquence à mémoire persistente
  optimale pour ce modèle.

  Cet article décrit également une implémentation dans PyTorch qui
  automatise le processus, peut être utilisée avec un RNN séquentiel
  quelconque dont les couches internes peuvent être arbitrairement
  complexes, et l'exécute en suivant la stratégie optimale étant
  donnée une limite de mémoire. À travers de nombreuses expériences,
  nous montrons que notre implémentation obtient invariablement de
  meilleures performances que les approches existantes sur une large
  gamme de réseaux, tailles d'images et tailles de batch.}

\begin{document}

\makeRR

\section{Introduction}
\label{sec.intro}
Training Deep Neural Network (DNN) is a memory-intensive
operation. Indeed, the training algorithms of most DNNs require to
store both the model weights and the forward activations in order to
perform back-propagation. In practice, training is performed
automatically and transparently to the user through autograd tools for
back-propagation, such as {\tt tf.GradientTape} in TensorFlow or {\tt
  torch.autograd.backward} in PyTorch. Unfortunately, the memory
limitation of current hardware often prevents data scientists from
considering larger models, larger image sizes or larger batch sizes
\cite{rota2018place,pleiss2017memory}.  This becomes even more
critical when learning has to be performed onto a low memory device as
it happens in a growing number of IoT applications
\cite{verhelst2017embedded}. On the other hand, model
parallelism~\cite{modelp} can be used to distribute, share and balance
the weights and the activations onto potentially distributed memory
nodes. The memory reduction strategy that we propose in this paper can
be applied to either a centralized setting or to each individual node
of a distributed setting. It consists in modifying autograd tools in
order to find a sequence of forward and backward operations, longer
than the sequence automatically performed by the autograd tools, but
for which it is possible to finely control memory consumption and thus
to adapt to the capabilities of the devices.

Memory consumption has been considered for a long time in the
framework of Automatic Differentiation
(AD)~\cite{griewank2008evaluating}. For a given batch size and a given
network model and even on a single node without relying on model
parallelism strategies, it enables to save memory at the price of
recomputations of forward activations. In the context of classical AD,
networks can be seen as (long) homogeneous (i.e., all stages are
identical) chains, and the forward activation corresponding to the
$i-$th stage of the chain has to be kept into memory until the
associated $i-$th backward stage. Checkpointing strategies are needed
to determine in advance which forward checkpoints should be kept into
memory and which should be recomputed from stored checkpoints during
the execution of the backward phase. Several studies have been
performed to determine optimal checkpointing strategies for AD in
different contexts, both in the case of homogeneous chains where
closed form formulas have been proposed~\cite{walther2004advantages},
and in the case of heterogeneous computation times, where Dynamic
Programming provides optimal solutions~\cite{griewank2008evaluating}
thanks to the \emph{memory persistency} property of all optimal solutions. Results on
homogeneous chains have been translated for specific DNNs such as
Recurrent Neural Networks
(RNNs)~\cite{gruslys2016memory}. Independently, a simple checkpointing
approach~\cite{periodiccheckpointing} has been proposed and is
available in PyTorch, based on a (non-optimal) strategy that involves
a sublinear number of checkpoints~\cite{chen2016training}.  In the
present paper, we propose several improvements and generalizations of
these results.

The main contribution of this paper is a careful modeling, presented
in Section~\ref{sec.model}, of the checkpointing operations that are
available in DNN frameworks. We show that autograd tools offer more
general operations and thus more optimization opportunities than those
used in the AD literature.  We assume that the DNN is given as a
linear sequence of modules, where internal modules can be arbitrarily
complex. In practice, this assumption does not hinder the class of
models that can be considered, and we propose implementations of
classical networks (ResNet, Inception, VGG, DenseNet) under this
model. In Section~\ref{sec.algo}, we show that models with
heterogeneous activation sizes (in addition to the heterogeneous computation
times that have already been considered in the literature), 
no longer satisfy the memory
persistency property, and we derive an algorithm to obtain optimal
memory persistent solutions. We show through an extensive experimental
evaluation in Section~\ref{sec.implem} that these additional
operations indeed enable to significantly increase the throughput (the
average number of processed images per second) when performing
training.

Another contribution of this paper is a complete and easy-to-use
implementation of the algorithm we propose in the PyTorch framework,
which is described in Section~\ref{sec.implem}. This tool
automatically measures the characteristics (memory consumption,
computation time) of each layer of the DNN, and computes the forward
and the backward phases while enforcing a memory limit, at the cost of
a minimal amount of recomputations. Therefore, we provide both new
original theoretical results that generalize the results achieved by
AD literature to a much larger class of models and operations, and we
propose a fully automatic tool that runs a mini-batch training
strategy while enforcing a memory constraint.

Note that throughout this paper, our goal is to propose a schedule and
a memory management strategy that enables to use less memory, but that
computes exactly the same results, at the price of some extra
computations. Therefore, the training strategy that we propose is
completely orthogonal to the optimization of the hyper-parameters of
the DNN: it will provide exactly the same accuracy after the same
number of epochs, at the benefit of a (much) lower memory consumption and at
the price of a (slightly) higher completion time.

\section{Related Work}
\label{sec.related}
%% LED: Missing references, and far from us : I remove it to gain
%% space. Put it back if you think it is important.

%% A first line of research is rather
%% concerned by memory issues during the inference phase, typically for
%% running previously trained networks on low-memory devices such as
%% smartphones or edges. Strategies such as network lightening \cite{},
%% network compression \cite{chen2018constraint} or knowledge distilling
%% \cite{} are used to reduce the memory footprint of the model once it
%% has been trained, usually at the price of a loss in accuracy
%% (typically around 1\% or 2\%).

Memory consumption is becoming an important issue in deep learning
today and covers several different aspects. In this paper, we focus on
memory issues at training time. A line of research for this purpose
consists in designing and training memory efficient architectures and
attempting to reach the same performance as state-of-the-art
networks. Reversible neural
networks~\cite{gomez2017reversible,chang2018reversible} (RevNet), for
instance, allow by design to run the back-propagation algorithm
without storing the forward activations.  Quantized neural networks
\cite{rastegari2016xnor,hubara2017quantized} rather try to reduce the
memory consumption by turning the network weights and/or activations
into binary or quantized variables. Other \textit{ad-hoc}
architectures such as MobileNets~\cite{howard2017mobilenets} or
ShuffleNet \cite{zhang2018shufflenet} finally try to sparsify the
network architecture so as to reduce the model size. In this paper,
however, we rather consider methods that reduce the memory footprint
of a given fixed model or architecture, while obtaining the exact same
output. Within this line of research, different categories of methods
like activation recomputation or layer optimization can be considered.
   
Recomputation is applied more and more to reduce memory. For example,
the authors of~\cite{pleiss2017memory} show, for a popular neural
network like DenseNet, that using shared memory storages and
recomputing concatenation and batch normalisation operations during
back-propagation help to go from quadratic memory cost to linear
memory cost for storing feature maps. Along the same idea,
re-implementations of some commonly used layers like batch
normalisation has been proposed~\cite{rota2018place}. In the latter
case, memory usage has been reduced by rewriting the gradient
calculation for this layer so that it does not depend on certain
activation values (so that it is no longer necessary to store
them). As mentioned in the introduction, model parallelism approach
has been advocated in many papers~\cite{modelp} and it can be combined
with data parallelism~\cite{datap}. Another
solution~\cite{shriram,rhu2016vdnn} is to offload some of the
activations from the memory of the GPU to the memory of the CPU, and
then to bring them back when they are needed during the backward
phase. Finally, Domain Decomposition or Spatial parallelism techniques
can be used to limit the memory required for storing forward
activations. In~\cite{Nikoli}, splitting large images into smaller
images allows to train in parallel the network on the small images
(augmented by a halo), at the price of extra communications in order
to synchronize parameter updates.  Both activation offloading and
spatial parallelism approaches are orthogonal to our approach and they
could be combined in order to achieve larger savings. We concentrate
in the present paper on the strategy that consists in recomputing
forward activation and we leave the combination of these approaches
(model parallelism, activation offloading and domain decomposition)
for future work.

When the network is a single chain of layers, the computation of the
gradient descent in the training phase is similar to Automatic
Differentiation (AD). The computation of adjoints has always been a
trade-off between recomputations and memory requirements and the use
of checkpointing strategies in the context of AD has been widely
studied.  Many studies have been performed to determine optimal
checkpointing strategies for AD in different contexts, depending on
the presence of a single or multi level
memory~\cite{aupy2016optimal}. Closed form formulas providing the
exact position of checkpoints have even been
proposed~\cite{griewank2000algorithm} for homogeneous chains (where
all layers are identical). When computation times are heterogeneous,
but activation sizes are identical, an optimal checkpointing strategy
can be obtained with Dynamic
Programming~\cite{griewank2008evaluating}. A generic
divide-and-conquer approach based on compiler techniques allows to
perform automatic differentiation for arbitrary
programs~\cite{siskind2018divide}.

The use of checkpointing strategies has recently been advocated for
Deep Neural Network (DNN) in several papers. A direct adaptation of
the results on homogeneous chains was proposed for the case of
Recurrent Neural Networks (RNNs)~\cite{gruslys2016memory}, but can not
extend to other DNNs. In an appendix to this work, a dynamic
programming formulation is given to solve the fully heterogeneous
problem (where both computation times and activation sizes of all
layers can be different). This formulation is close to the work
presented here, but is restricted to checkpointing only the layer
outputs, and no implementation is provided. Another generalization of
the result on homogeneous chains allows to obtain optimal
checkpointing strategies for join
networks~\cite{beaumont2019multichain}, which are made of several
homogenenous chains joined together at the end.

On the other hand, an implementation of checkpointing exists in
PyTorch~\cite{periodiccheckpointing}, based on a simple periodic
checkpointing strategy which exploits the idea presented
in~\cite{chen2016training}. In this strategy, the chain is divided in
equal-length segments, and only the input of each segment is stored
during the forward phase. This strategy provides non-optimal solutions
in terms of throughput and memory usage, because it does not benefit
from the fact that more memory is available when computing the
backward phase of the first segment (since values stored for later
segments have already been used). This implementation was used to
be able to process significantly larger models~\cite{memonger}. 

To the best of our knowledge, this work is the first attempt to
precisely model heterogeneity and more importantly the ability,
offered in DNN frameworks, to combine two types of activation savings,
by either storing only the layer inputs (as done in AD literature), or
by recording the complete history of operations that produced the
outputs (as available in autograd tools), as described in
Section~\ref{sec.model}.

\section{Modeling and Problem Formulation}
\label{sec.model}

We present here the computation model used throughout the
paper to describe the different checkpointing strategies that can be
used during an iteration of the back-propagation algorithm. We also
highlight how this model differs from the classic Automatic
Differentiation model~\cite{griewank1989automatic}. 

\subsection{Model for the Back-propagation Algorithm}
\begin{figure}%[htb]
  \tikzset{fwopG/.style={rectangle, thick, draw=black, fill=black!20, font=\footnotesize}}
  \tikzset{bwopG/.style={fwopG}}

  \begin{subfigure}[b]{\columnwidth}
    \begin{center}
      \resizebox{\columnwidth}{!}{
	\begin{tikzpicture}[>=latex]
	  % The various elements are conveniently placed using a matrix:
	  \matrix[row sep=0.9cm,column sep=1.0cm,ampersand replacement=\&] {
	    % First line: Forward operations
	    %\&
	    \node (finput)  [] {};	\&
	    \node (f_1) [fwopG]{$F^1$};	\&
	    \node (f_2)   [fwopG]{$F^2$};     \&
	    \node (f_x)   {$\cdots$};     \&
	    \node (f_l-2) [fwopG]{$F^{L-1}$}; \&
	    \node (f_l-1) [fwopG]{$F^{L}$}; \&
	    \node (f_l) [fwopG]{$F^{L+1}$}; \&
	    \node (foutput) [] {};\&
	    \\
	    % Second line: Backward operations
	    \node (boutput) [] {};	\&
	    \node (b_2l+1) [bwopG]{$B^1$};	\&
	    \node (b_2l) [bwopG]{$B^2$};	\&
	    \node (b_2l-1)   [bwopG]{$B^3$};     \&
	    \node (b_x)   {$\cdots$};     \&
	    \node (b_l+1) [bwopG]{$B^{L}$}; \&
	    \node (b_l+2) [bwopG]{$B^{L+1}$}; \&
	    \node (binput) [] {};	\&
	    \\
	  };
	  
	  % The diagram elements are now connected through arrows:
	  \path[->]
	  % Forward path Ti -> Ti+1
	  (finput) edge node [pos=0.5, above] {$a^0$} (f_1)
	  (f_1) edge node [pos=0.5, above] {$a^1$} (f_2)
	  (f_2) edge node [pos=0.5, above] {$a^2$} (f_x)
	  (f_x) edge node [pos=0.5, above] {$a^{L-2}$} (f_l-2)
	  (f_l-2) edge node [pos=0.5, above] {$a^{L-1}$} (f_l-1)
	  (f_l-1) edge node [pos=0.5, above] {$a^{L}$} (f_l)
	  (f_l) edge node [pos=0.5, above] {$loss$} (foutput)
	  % Backward path Ti -> Ti+1
	  (binput) edge node [pos=0, below] {$\delta^{L+1} = 1$} (b_l+2)
	  (b_l+2) edge node [pos=0.5, below] {$\delta^{L}$} (b_l+1)
	  (b_l+1) edge node [pos=0.5, below] {$\delta^{L-1}$} (b_x)
	  (b_x) edge node [pos=0.5, below] {$\delta^{3}$} (b_2l-1)
	  (b_2l-1) edge node [pos=0.5, below] {$\delta^{2}$} (b_2l)
	  (b_2l) edge node [pos=0.5, below] {$\delta^{1}$} (b_2l+1)
	  (b_2l+1) edge node [pos=0.5, below] {$\delta^{0}$} (boutput)
	  
	  % Transversal paths
	  (finput) edge node [pos=0.4, right] {$\;a^0$} (b_2l+1)
	  (f_1) edge node [pos=0.4, right] {$\;a^1$} (b_2l)
	  (f_2)   edge node [pos=0.4, right] {$\;a^2$} (b_2l-1)
	  (f_l-2)   edge node [pos=0.4, right] {$\;\;a^{L-1}$} (b_l+1)
	  (f_l-1)   edge node [pos=0.4, right] {$\;\;a^{L}$} (b_l+2)
	  
	  % Down paths
	  (f_1) edge node [pos=0.55, right] {$\bar{a}^1$} (b_2l+1)
	  (f_2) edge node [pos=0.55, right] {$\bar{a}^2$} (b_2l)
	  (f_x) edge node [pos=0.55, right] {$\bar{a}^3$} (b_2l-1)
	  (f_l-2) edge node [pos=0.55, right] {$\bar{a}^{L-1}$} (b_x)
	  (f_l-1) edge node [pos=0.55, right] {$\bar{a}^{L}$} (b_l+1)
	  (f_l) edge node [pos = 0.55, right] {$loss$} (b_l+2)
	  ;
	  
	\end{tikzpicture}
      }
    \end{center}

    \caption{Graph for a general sequential deep neural network.}

    \label{fig:PyTorch.adj}
  \end{subfigure}

  \begin{subfigure}[b]{\columnwidth}
    \begin{center}
      \resizebox{\columnwidth}{!}{
	\begin{tikzpicture}[>=latex]
	  % The various elements are conveniently placed using a matrix:
	  \matrix[row sep=0.9cm,column sep=1.0cm,ampersand replacement=\&] {
	    % First line: Forward operations
	    %\&
	    \node (finput)  [] {};	\&
	    \node (f_1) [fwopG]{$F^1$};	\&
	    \node (f_2)   [fwopG]{$F^2$};     \&
	    \node (f_x)   {$\cdots$};     \&
	    \node (f_l-2) [fwopG]{$F^{L-1}$}; \&
	    \node (f_l-1) [fwopG]{$F^{L}$}; \&
	    \node (f_l) [fwopG]{$F^{L+1}$}; \&
	    \node (foutput) [] {};\&
	    \\
	    % Second line: Backward operations
	    \node (boutput) [] {};	\&
	    \node (b_2l+1) [bwopG]{$B^1$};	\&
	    \node (b_2l) [bwopG]{$B^2$};	\&
	    \node (b_2l-1)   [bwopG]{$B^3$};     \&
	    \node (b_x)   {$\cdots$};     \&
	    \node (b_l+1) [bwopG]{$B^{L}$}; \&
	    \node (b_l+2) [bwopG]{$B^{L+1}$}; \&
	    \node (binput) [] {};	\&
	    \\
	  };
	  
	  % The diagram elements are now connected through arrows:
	  \path[->]
	  % Forward path Ti -> Ti+1
	  (finput) edge node [pos=0.5, above] {$a^0$} (f_1)
	  (f_1) edge node [pos=0.5, above] {$a^1$} (f_2)
	  (f_2) edge node [pos=0.5, above] {$a^2$} (f_x)
	  (f_x) edge node [pos=0.5, above] {$a^{L-2}$} (f_l-2)
	  (f_l-2) edge node [pos=0.5, above] {$a^{L-1}$} (f_l-1)
	  (f_l-1) edge node [pos=0.5, above] {$a^{L}$} (f_l)
	  (f_l) edge node [pos=0.5, above] {$loss$} (foutput)
	  % Backward path Ti -> Ti+1
	  (binput) edge node [pos=0, below] {$\delta^{L+1} = 1$} (b_l+2)
	  (b_l+2) edge node [pos=0.5, below] {$\delta^{L}$} (b_l+1)
	  (b_l+1) edge node [pos=0.5, below] {$\delta^{L-1}$} (b_x)
	  (b_x) edge node [pos=0.5, below] {$\delta^{3}$} (b_2l-1)
	  (b_2l-1) edge node [pos=0.5, below] {$\delta^{2}$} (b_2l)
	  (b_2l) edge node [pos=0.5, below] {$\delta^{1}$} (b_2l+1)
	  (b_2l+1) edge node [pos=0.5, below] {$\delta^{0}$} (boutput)
	  
	  % Transversal paths
	  (finput) edge node [pos=0.4, right] {$\;a^0$} (b_2l+1)
	  (f_1) edge node [pos=0.4, right] {$\;a^1$} (b_2l)
	  (f_2)   edge node [pos=0.4, right] {$\;a^2$} (b_2l-1)
	  (f_l-2)   edge node [pos=0.4, right] {$\;\;a^{L-1}$} (b_l+1)
	  (f_l-1)   edge node [pos=0.4, right] {$\;\;a^{L}$} (b_l+2)
	  
	  % Down paths
	  (f_l) edge node [pos = 0.55, right] {$loss$} (b_l+2)
	  ;
	  
	  %    \begin{pgfonlayer}{background}
	  %        \node [background, fit=(input) (b_l+1)] {};
	  %    \end{pgfonlayer}
	\end{tikzpicture}
      }
    \end{center}
    
    \caption{Graph for an automatic differentiation application.}
    \label{fig:ad.adj}
  \end{subfigure}
  \caption{Graphs of a general sequential Deep Neural Network and an Automatic Differentiation application.
  }
  \label{fig:adj}
\end{figure}
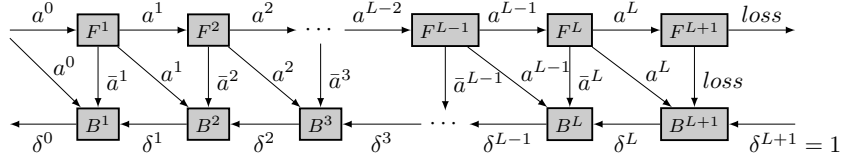
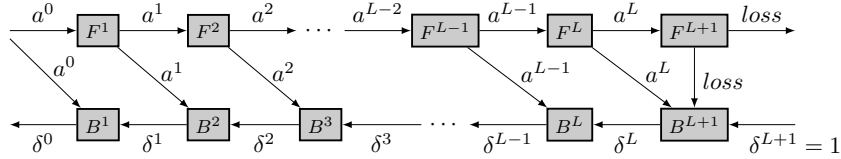

We consider a chain of $L$ stages (\textit{i.e} layers or blocks of
layers), numbered from $1$ to $L$. Each stage $\ell$ is associated to
a forward operation $F^{\ell}$ and a backward operation $B^{\ell}$
(see Figure~\ref{fig:PyTorch.adj}). For notational convenience, the
computation operations of the loss $\mathcal{L}$ are denoted $F^{L+1}$
and $B^{L+1}$. We denote by $a^{\ell}$ the activation tensor output of
$F^{\ell}$ and by $\delta^{\ell}=\frac{\partial \mathcal{L}}{\partial
  a^{\ell}}$ the back-propagated intermediate value provided as input
of the backward operation $B^{\ell}$. For a simple Fully Connected
(FC) layer, we would have the following forward and back-propagation
equations:

\begin{align*}
F^{\ell}: & \;a^{\ell}=\sigma(w^{\ell} a^{\ell-1} + b^{\ell})\\
B^{\ell}: & \;\delta^{\ell-1}= {(w^{\ell})}^T ( \delta^{\ell} \odot  \sigma’(z^{\ell}) )\\
  & \;\frac{\partial \mathcal{L}}{\partial w^{\ell}}=a^{\ell-1}(\delta^{\ell}\odot\sigma'(z^\ell))\\
& \;\frac{\partial \mathcal{L}}{\partial b^{\ell}}=\delta^{\ell}\odot\sigma'(z^\ell)
\end{align*}

where $w^{\ell}$ and $b^{\ell}$ are the parameters of the FC layer to
be learned, $\sigma$ is the non linear activation function and
$z^{\ell}$ is the pre-activation vector (\textit{i.e.}
$a^{\ell}=\sigma(z^{\ell})$). For complex blocks of layers
(\textit{e.g.} inception modules or residual blocks), $F^{\ell}$ and
$B^{\ell}$ are more complex functions that can be expressed as

\begin{align*}
F^{\ell}:\;& a^{\ell}=f_{\ell}(\theta^{\ell},a^{\ell-1})\\
B^{\ell}:\;& \delta^{\ell-1}= \bar{f}_{\ell}(\theta^{\ell},\delta^{\ell},\bar{a}^{\ell}, a^{\ell -1})\\
& \;\frac{\partial \mathcal{L}}{\partial \theta^{\ell}}=\bar{g}_{\ell}(\delta^{\ell},\bar{a}^{\ell}, a^{\ell -1}),
\end{align*}

where $\theta^{\ell}$ is the whole set of parameters of the block and
$\bar{a}^{\ell}$ is the set of all intermediate activation values that
are required to compute the back-propagation inside the block,
including $a^{\ell}$ but not including $a^{\ell-1}$ (in the simple
case of the FC layer we have $\bar{a}^{\ell}=\{a^{\ell},
z^{\ell}\}$). In classical implementations of the back-propagation
algorithm, all activation values are stored in memory during the
forward step $F^{\ell}$ until the backward step $B^{\ell}$ is
completed (in practice, for most frameworks, the full computational
graph allowing to compute $a^{\ell}$ is stored).

The principle of checkpointing is to trade memory for computing time
by not saving all activations in memory but recomputing them when
needed by the backward steps. Therefore, let us introduce three
different types of forward operations: (i) \nforop[\ell] allowing to
compute $F^{\ell}$ without saving any data in memory, (ii)
\cforop[\ell] allowing to compute $F^{\ell}$ while saving the input
$a^{\ell-1}$ of the block (\textit{i.e.} \textit{checkpointing}) and
(iii) \eforop[\ell] allowing to compute $F^{\ell}$ while saving all
the intermediate data $\bar{a}^{\ell}$ required by the backward
step. Note that $B^{\ell}$ cannot be computed until $F_{all}^{\ell}$
has been processed. However, $F_{all}^{\ell}$ uses more memory than
\cforop[\ell], so it may be more efficient to compute \cforop[\ell]
first and then compute $F_{all}^{\ell}$ from $a^{\ell-1}$ later in the
sequence of instructions. Overall, the problem is to find the optimal
sequence of operations that minimizes the computation time while
taking into account the memory constraint.

In the following, we assume that the memory needed to store each data
item is known (Section~\ref{sec.implem} describes how this information
can be measured automatically before starting the actual training of
the model). We denote as $\wx[\ell]$ the memory required to store
$a^{\ell}$, $\wbx[\ell]$ to store $\bar{a}^{\ell}$ and $\wy[\ell]$ to
store $\delta^{\ell}$ (in practice, $\wx[\ell]=\wy[\ell]$). Note that
we focus here on the memory used by activations. We assume that the
memory required to store the model and the gradients of the model
parameters has already been allocated and removed from the available
memory.
  
A strategy for computing $\delta_0$ given $z_0$ is a sequence of
operations, the list of which is described in
Table~\ref{operationsPyTorch}. Each operation requires a certain input
and produces a certain output, which replaces the input in memory.

Since each stage in the chain can be arbitrarily complex, it may
have a memory peak higher than the sum of its input and output data.
This is modeled by introducing the memory overhead of operations: we assume that
the memory needed to compute an operation is the sum of its
input, output data and memory overhead.

We consider that, at the beginning, the memory contains $\{a^0=x\}$,
\textit{i.e.} the input data.  The processing of a sequence consists
in executing all the operations one after the other, replacing the
inputs of each operation by its outputs in the memory. The sequence is
called \emph{valid} if for any operation, its input is present in the
memory when processed. For example, for $L=4$, a possible valid
sequence for the computation is:

$$\cforop[1], \nforop[2], \cforop[3],
  \eforop[4], \eforop[5], B^5, B^4, \eforop[3], B^3, \eforop[1],
  \eforop[2], B^2, B^1$$

The maximum memory usage of a valid sequence is the maximum, for all
operations, of the size of the data in memory during the operation,
plus the peak usage of this operation. The computation time of a
sequence is the sum of the durations of its operations. 
The optimization problem is thus, given a memory limit $M$,
to find a valid sequence with a memory usage not exceeding $M$ and
whose computation time is minimal.

\begin{table*}[h]
  \footnotesize
  \begin{center}
    \begin{tabular}[h]{|c l |c|c|c|c|}
      \hline
      & Operation & Input & Output & Time & Memory overhead \\
      \hline
      \multirow{2}{*}{$\eforop[\ell]$} & \multirow{2}{*}{Forward and save all} & $\{a^{\ell-1}\}$ & $\{a^{\ell-1},~ \bar{a}^{\ell}\}$ &
      \multirow{2}{*}{$\uf[\ell]$} & \multirow{2}{*}{$\of[\ell]$} \\
      &  & $\{\bar{a}^{\ell-1}\}$ & $\{\bar{a}^{\ell-1},~ \bar{a}^{\ell}\}$ & &\\  
      \hline
      \multirow{2}{*}{$\cforop[\ell]$} & \multirow{2}{*}{Forward and
        checkpoint input} & $\{a^{\ell-1}\}$ & $\{a^{\ell-1},~ a^{\ell}\}$ &
      \multirow{2}{*}{$\uf[\ell]$} & \multirow{2}{*}{$\of[\ell]$}\\
      &  & $\{\bar{a}^{\ell-1}\}$ & $\{\bar{a}^{\ell-1},~ a^{\ell}\}$ & &\\
      \hline
      $\nforop[\ell]$ & Forward without saving & $\{a^{\ell-1}\}$ &
      $\{a^{\ell}\}$ & $\uf[\ell]$ & $\of[\ell]$ \\
      \hline
      \multirow{2}{*}{$B^l$} & \multirow{2}{*}{Backward
        step} & $\{\delta^{\ell},~ \bar{a}^{\ell},~ a^{\ell-1}\}$ & $\{\delta^{\ell -1}\}$ &
      \multirow{2}{*}{$\ub[\ell]$} & \multirow{2}{*}{$\ob[\ell]$} \\
      &  & $\{\delta^{\ell},~ \bar{a}^{\ell},~ \bar{a}^{\ell-1}\}$ & $\{\delta^{\ell -1},~ \bar{a}^{\ell-1}\}$ & &\\
      \hline
    \end{tabular}
  \end{center}
  
  \caption{Operations performed by a schedule. The second line
    shows the behavior when $\bar{a}^{\ell-1}$ is used instead of
    $a^{\ell-1}$.}
  \label{operationsPyTorch}
\end{table*}

\subsection{Difference with Automatic Differentiation Models}
In the context of Automatic Differentiation (AD), the computational
graph has a similar structure (see Figure~\ref{fig:ad.adj}).  The main
difference comes from the absence of $\bar{a}$ dependencies between a
forward operation and the corresponding backward operation.  In AD,
backward operations also require the intermediate activation values
but, in general, forward computations are recomputed using a special
mode called \emph{taping}, that stores intermediate activation values
right before processing the corresponding backward
operation~\cite{griewank1989automatic}.  Several consecutive forward
operations can be taped to execute the corresponding backward
operations successively, which is equivalent to considering these
forward operations as one big forward meta-transaction.  Nevertheless,
to the best of our knowledge, there has been no study on models
allowing taping forward operations during the forward phase for later
usage during the backward phase.  Our more relaxed model allows more
freedom (and thus higher efficiency, as seen in
Section~\ref{sec.implem}), since each forward operation can be taped
(using a $\eforop[\ell]$ operation as stated above) even if the
corresponding backward operation is not executed immediately after it.
Optimal solutions for chains with heterogeneous computing time in the
automatic differentiation model are
known~\cite{griewank2008evaluating}. As we show in the next Section,
considering heterogeneous activation sizes makes it more difficult to
obtain optimal solutions, since the property of \emph{memory
  persistency} no longer holds. A dynamic program to compute the best
memory persistent solution for heterogeneous chains in the automatic
differentiation model was proposed
in~\cite{gruslys2016memory}. However, the optimal execution of a
computation graph for a deep neural network
(Figure~\ref{fig:PyTorch.adj}) cannot be directly derived from the
optimal solution for the Automatic Differentiation case
(Figure~\ref{fig:ad.adj}) and a deeper analysis of the problem is
required, which is the main contribution of the
Section~\ref{sec.algo}.

\section{Optimal Checkpointing Algorithm}
\label{sec.algo}

In this Section, we analyze the problem defined above. We first
present the memory persistency property, used in optimality proofs of
dynamic programming algorithms in the automatic differentiation
literature. We show that models with heterogeneous activation sizes do
not satisfy the memory persistency property. This applies to the model
presented above, and the automatic differentiation model shown on
Figure~\ref{fig:ad.adj} and used in~\cite{gruslys2016memory}. However,
finding optimal non persistent solutions appear to be a difficult
challenge, so we focus on obtaining the best persistent strategy, for
which we can derive an optimal dynamic programming algorithm.

\subsection{Considerations on Memory Persistency}

A schedule is said to be \emph{memory
  persistent}~\cite{griewank2008evaluating} if any checkpointed value
is kept in memory until it is used in the backward phase. A key
observation for homogeneous activation sizes is that all optimal
schedules are memory persistent: if an activation $a_i$ is
checkpointed, but deleted before being used for $B^{i+1}$, it is
actually more efficient to checkpoint $a_{i+1}$ since it avoids to
recompute $F^{i+1}$.

\begin{figure}
  \tikzset{fwopG/.style={rectangle, thick, draw=black, fill=black!20, font=\footnotesize}}
  \tikzset{bwopG/.style={fwopG}}
  \begin{center}
    \resizebox{\columnwidth}{!}{
      \begin{tikzpicture}[>=latex]
	% The various elements are conveniently placed using a matrix:
	\matrix[row sep=0.9cm,column sep=1.0cm,ampersand replacement=\&] {
	  % First line: Forward operations
	  %\&
	  \node (finput)  [] {};	\&
	  \node (f_1) [fwopG]{$k$};	\&
	  \node (f_2)   [fwopG]{$2$};     \&
          \node (f_3)   [fwopG]{$0$}; \&
	  \node (f_x)   {$\cdots$};     \&
	  \node (f_l-2) [fwopG]{$0$}; \&
	  \node (f_l-1) [fwopG]{$0$}; \&
	  \node (f_l) [fwopG]{$0$}; \&
	  \node (foutput) [] {};\&
	  \\
	  % Second line: Backward operations
	  \node (boutput) [] {};	\&
	  \node (b_2l+1) [bwopG]{$0$};	\&
	  \node (b_2l) [bwopG]{$0$};	\&
	  \node (b_2l-1)   [bwopG]{$0$};     \&
	  \node (b_2l-2)   [bwopG]{$0$};     \&
	  \node (b_x)   {$\cdots$};     \&
	  \node (b_l+1) [bwopG]{$0$}; \&
	  \node (b_l+2) [bwopG]{$0$}; \&
	  \node (binput) [] {};	\&
	  \\
	};
	
	% The diagram elements are now connected through arrows:
	\path[->]
	% Forward path Ti -> Ti+1
	(finput) edge node [pos=0.5, above] {$0$} (f_1)
	(f_1) edge node [pos=0.5, above] {$1$} (f_2)
	(f_2) edge node [pos=0.5, above] {$2$} (f_3)
	(f_3) edge node [pos=0.5, above] {$3$} (f_x)
	(f_x) edge node [pos=0.5, above] {$3$} (f_l-2)
	(f_l-2) edge node [pos=0.5, above] {$3$} (f_l-1)
	(f_l-1) edge node [pos=0.5, above] {$4$} (f_l)
	(f_l) edge node [pos=0.5, above] {$0$} (foutput)
	% Backward path Ti -> Ti+1
	(binput) edge node [pos=0, below] {$0$} (b_l+2)
	(b_l+2) edge node [pos=0.5, below] {$0$} (b_l+1)
	(b_l+1) edge node [pos=0.5, below] {$0$} (b_x)
	(b_x) edge node [pos=0.5, below] {$0$} (b_2l-2)
	(b_2l-2) edge node [pos=0.5, below] {$0$} (b_2l-1)
	(b_2l-1) edge node [pos=0.5, below] {$0$} (b_2l)
	(b_2l) edge node [pos=0.5, below] {$0$} (b_2l+1)
	(b_2l+1) edge node [pos=0.5, below] {$0$} (boutput)
	
	% Transversal paths
	(finput) edge node [pos=0.4, right] {$0$} (b_2l+1)
	(f_1) edge node [pos=0.4, right] {$1$} (b_2l)
	(f_2)   edge node [pos=0.4, right] {$2$} (b_2l-1)
	(f_3)   edge node [pos=0.4, right] {$3$} (b_2l-2)
	(f_l-2)   edge node [pos=0.4, right] {$3$} (b_l+1)
	(f_l-1)   edge node [pos=0.4, right] {$4$} (b_l+2)
	
	% Down paths
	(f_l) edge node [pos = 0.55, right] {$0$} (b_l+2)
	;

        \draw[snake=brace, raise snake=0.2cm, segment
          amplitude=0.25cm] (f_3.north west) -- (f_l-1.north east) node[midway,above=0.45cm]{$n$}; 
        
	%    \begin{pgfonlayer}{background}
	%        \node [background, fit=(input) (b_l+1)] {};
	%    \end{pgfonlayer}
      \end{tikzpicture}
    }
  \end{center}
  
  \caption{Counter example where no memory persistent solution is
    optimal. Values on the edges represent the size of the
    activations, values inside nodes represent the computing time of
    the layers. The memory limit is $M=8$.}
  \label{fig:persistent.counter.example}
\end{figure}
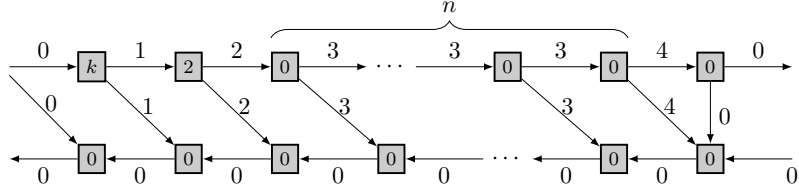

However, when activation sizes are heterogeneous, this property no
longer holds. We show an example on
Figure~\ref{fig:persistent.counter.example}: the chain length is $L =
n+2$ for any $n$, all backward sizes $\wy[\ell]$ and computing times
$\ub[\ell]$ are $0$, as well as most of the forward computation times,
except $\uf[1] = k$ and $\uf[2] = 2$. Most forward activations sizes
$\wx[\ell]$ are $1$, except $\wx[2] = \wx[L] = 2$. The memory limit is
$M=8$.

Since computing $F^L$ requires a memory of $7$, it is not possible to
checkpoint $a^2$ (whose size is $2$) in the forward phase. We can thus
identify two valid memory persistent schedules, which are candidates
for optimality: either $a^{1}$ is checkpointed during the forward
phase, or it is not checkpointed. In the first case, $a^2$ is never
checkpointed, and thus $\forop[2]$ is processed $n+1$ times. This
results in a makespan $T_1 = k + 2(n + 1)$. In the second case, the
forward phase is performed with only $a^0$ stored in memory, until
$B^{L}$ is computed. Then, the computation starts from the beginning,
and this time it is possible to checkpoint $a^2$, which allows to
compute all of the 0-cost \forop[\ell] without recomputing $F^2$. At
the end it is necessary to recompute $F^1$, which results in a makespan
$T_2 = 2(k+2) + k = 3k + 4$.

It is also possible to imagine the following non-persistent schedule:
$a^1$ is checkpointed during the forward phase, and kept in memory
until the second time that $F^2$ is computed. Indeed, at that time
$F^L$ has already been computed, and it is possible to checkpoint
$a^2$ instead of $a^1$ (but not both at the same time since computing $F^{L-1}$ requires a memory of $6$). At the end it
is necessary to recompute $F^1$, and this results in a makespan $T_0 =
k + 2\times2 + k = 2k + 4$.

Setting $k = n-1$ ensures that $T_1 = T_2 = 3n + 1$, while $T_0 =
2n+2$. In that case, the makespan of the non-persistent schedule is
thus lower than the makespan of any memory persistent schedule.
Nevertheless, as a heuristic to the general problem, in the rest of
the paper we search for memory persistent schedules. We obtain an
optimal algorithm in the next Section, and show in experimental
evaluation that this allows to obtain significant improvement over
existing solutions. 

\subsection{Optimal Persistent Schedule}
In this section, we present an algorithm based on Dynamic Programming
to obtain the optimal persistent schedule. For a chain of length $L$, we
denote by $\optBP$ the 
optimal execution time to process the chain from stage $s$ to stage $t$
with peak memory at most $m$, assuming that the input tensors
$a^{s-1}$ and $\delta^t$ are stored in memory, but the size of
$a^{s-1}$ should not be counted in the memory limit $m$. Let us introduce
the following notations
\begin{align*} \nomem[s, t] &=\max \begin{cases}
    \wy[t] + \wx[s] + \of[s],  \\%& \text{first forward step} \\
    \wy[t] +\displaystyle \max_{s +1 \leq j < t} \Bigl\{\wx[j-1]  + \wx[j] + \of[j]\Bigr\}  %& \text{forward steps}\\ 
  \end{cases}\\ 
  \emem[s,t] &=  \max \begin{cases}
    \wy[t] + \wbx[s] + \of[s] , \\%& \text{ forward step}\\ 
    \wy[s]+  \wbx[s] + \ob[s] %& \text{backward step}\\ 
  \end{cases}
\end{align*}

$\nomem[s,t]$ for $1\leq s < t \leq L  +1$ denotes the memory peak to
compute all $\nforop$ steps from $s$ to $t$, and $\emem[s,t]$ for
$1\leq s \leq t \leq L +1$ denotes the memory peak to run $\eforop[s]$
and $B^s$.

\begin{theorem} \label{th:OptBP}
  $\optBP$, the optimal time for any valid persistent sequence
  to process the chain from stage $s$ to stage $t \geq s$ with available memory
  $m$, is given by
  \begin{align}
	\label{DP:Init}\optBP[s, s, m]  & = \begin{cases} \uf[s] + \ub[s] & m \geq \emem[s,s] \\ \infty & m < \emem[s,s] \end{cases}\\
    \optBP &= \min \left(C_1(s, t, m), \quad C_2(s, t, m) \right) \label{DP:mainrec} 
    \end{align}
    \begin{align*}
      C_1(s, t, m) &= \begin{cases} \displaystyle \min_{s' = s+1...t} C_{ck}(s, s', t, m) &  m \geq \nomem[s, t] \\ \infty & m < \nomem[s,t] \end{cases} \notag\\ 
      C_2(s, t, m) &= \begin{cases} C_{all} (s,t,m)  &  m \geq \emem[s, t]  \\
	\infty & m < \emem[s,t] 
      \end{cases} \notag  \text{,   where}\\
      %  \hspace{-0.7cm}$C_{ck}(s, s', t, m) = \displaystyle \sum_{k=s}^{s'-1} \uf[k] +  C_{\text{BP}}(s', t, m-\wx[s' -1]) +C_{\text{BP}}(s, s'-1, m)$\\
    \end{align*}
    \begin{align*}
      C_{ck}(s, s', t, m) & = 
      \begin{aligned}[t]
        \sum_{k=s}^{s'-1} \uf[k] & +  \optBP[{s', t, m-\wx[s' -1]}]\\& + \optBP[{s, s'-1, m}]
      \end{aligned}\\
      C_{all} (s,t,m) & = \uf[s]+ \optBP[{s+1, t, m - \wbx[s]}] + \ub[s]
    \end{align*}
\end{theorem}

We can interpret these values as follows: $C_{ck}(s, s', t, m)$ denotes
the computing time for the chain from $s$ to $t$ 
if forward operations from $s$ to $s' -1$ are processed with $\nforop$, 
whereas $a^{s-1}$ is stored in memory by $\cforop[s]$. 
$C_{all}(s, t, m)$ is the computing time for the chain 
from $s$ to $t$ if $F^s$ is processed with $\eforop[s]$. 

\begin{proof} 

We first start by showing that Eq.~\eqref{DP:Init} is a
correct initialisation of the dynamic programming. Indeed, in order to
back-propagate one layer, one needs to perform $\eforop[s]$ to be able
to execute $B^s$ afterwards. This requires a memory of size
$\emem[s,s]$: we consider that the size of the input of the chain $a^{s-1}$
is counted outside of the memory limit $m$, and $\emem[s,s]$ represents
the highest of the peak memory usage between forward and backward
operations corresponding to layer $s$.

Let us now provide the proof for the general case. Since we are
looking for a persistent schedule, and since the input tensor
$a^{s-1}$ is to be stored in memory, the optimal sequence has only two
possible ways to start: either with $\cforop[s]$ to store $a^{s-1}$
and compute $a^{s}$, or with $\eforop[s]$ to compute $\bar{a}^{s}$.

If the first operation is $\cforop[s]$, then we can denote $a^{s' -1}$ the
first value stored in memory after $a^{s-1}$ (since some $\eforop$
operation needs to be performed before the first backward, $a^{s'-1}$
necessarily exists). Due to memory persistence, and since while $a^{s' -1}$
is present in memory there is no need to consider any $a^k$ or
$\bar{a}^k$ for $s \leq k < s' -1$, the problem of computing $\delta_{s' -1}$ from
the input $a^{s'-1}$ is exactly the one corresponding to $\optBP[{s',t, m-\wx[s' -1]}]$. 
Indeed, we assume that $a^{s'-1}$ is to be stored in
memory, but count its memory usage outside the limit $m-\wx[s'-1]$. On
the other hand, once this chain is processed, the remaining part of
the chain represents another chain which starts at position $s$ and
finishes at $s' -1$, where the new currently stored gradient is $\delta_{s'-1}$ and
$a^{s' -1}$ is not needed anymore and thus is finally removed. Bringing
everything together yields the equation for $C_{ck}(s, s',  t, m)$. Choosing $s'$ 
so that it brings minimum of $C_{ck}(s, s', t, m)$ guarantees the best possible
 solution, which is reflected in $C_1(s,t,m)$.
 
If the first operation is $\eforop[s]$ then by definition the value
$\bar{a}_{s}$ will also be checkpointed. As memory persistence holds
and no other value $a^k$ or $\bar{a}^k$ for $0 \leq k \leq s-1$ is
needed until $B_{s+1}$, we see that the problem of computing $\delta^{s}$
is exactly the one corresponding to $\optBP[{s+1, t, m -
    \wbx[s]}]$, where the decrease in memory corresponds to the
memory needed to store $\bar{a}^{s}$.  After this chain is
completed, it is possible to perform the last backward step $B^s$ as
both $\bar{a}^{s}$ and $a^{s-1}$ are already stored. Provided that the
memory limits are not violated, we obtain the equation for $C_{all}(s, t, m)$.
 
At last, we show that the memory limits $\nomem[s,t]$ and
$\emem[s,t]$ are valid. The first one states that executing the
chain from $s$ to $t$ with $\delta^{t}$ stored requires at least
enough memory to execute all the forward steps without saving any activation.
 The second one states that executing the chain from $s$ to $t$ by starting with an
$\eforop[s]$ operation requires enough memory to perform this operation
with $\delta_{t}$ stored, and enough memory to perform the corresponding
backward operation.
\end{proof}

This theorem proves that Algorithm~\ref{alg:optBP} and
Algorithm~\ref{alg:optBP:rec} compute an optimal sequence, for all
input parameters. Indeed, the computing time of the returned sequence
is exactly $\optBP[1, L+1, M]$.

\begin{algorithm}
\caption{Compute optimal persistent schedule for a chain of length $L$ with memory $M$.}
\label{alg:optBP}
\begin{algorithmic}[1]
\State Initialize table $C$ of size $(L+1) \times (L+1)\times M$
\For {$1 \leq s\leq L+1$ \textbf{and} $1\leq m \leq M$}
\State Initialize $C[s,s,m]$ with equation~\eqref{DP:Init}
\EndFor
\For {$s = 1, \dots, L$}
  \For {$t = s +1, \dots, L+1$}
    \For {$m = 1, \dots, M$}
      \State Compute $C[s,t, m]$ with equation~\eqref{DP:mainrec}
    \EndFor 
  \EndFor
\EndFor
\State \Return $\text{OptRec}(C, 1, L+1, M-\wx[0])$ \Comment{Alg.~\ref{alg:optBP:rec}}
\end{algorithmic}
\end{algorithm}

\begin{algorithm}
  \caption{$\text{OptRec}(C, s, t, m)$ -- Obtain optimal persistent
    sequence from the table $C$}
\label{alg:optBP:rec}
\begin{algorithmic}
  \If{$C[s, t, m] = \infty$}
  \State \Return Infeasible
  \ElsIf{$s = t$}
  \State \Return $(\eforop[s], B^s)$
  \ElsIf{$C[s, t, m] = C_{ck}(s, s', t, m)$}
  \State $\mathcal{S} \gets (\cforop[s], \nforop[s+1], \dots, \nforop[s'])$
  \State $\mathcal{S} \gets (\mathcal{S}, \text{OptRec}(C, s', t, m - \wx[s'-1]))$
  \State \Return $(\mathcal{S}, \text{OptRec}(C, s, s'-1, m))$
  \Else
  \State \Return $(\eforop[s], \text{OptRec}(C, s+1, t, m - \wbx[s]), B^{s})$
  \EndIf
\end{algorithmic}

\end{algorithm}
\section{Implementation and Validation}
\label{sec.implem}

We demonstrate the applicability of our approach by presenting a tool
that allows the above algorithm to be used with any Pytorch DNN based
on the \texttt{nn.Sequential} container. This tool is used in a very
similar fashion to the existing \texttt{checkpoint\_sequential} tool
already available in PyTorch~\cite{periodiccheckpointing}, but
offers a much more optimized checkpoint selection.  Our tool works
in three phases: parameter estimation, optimal sequence computation
and sequence processing. It is expected that the first two phases are
performed only once, before the start of the training, while the sequence is
used at each iteration.

\subsection{Parameter Estimation}
In the parameter estimation phase, the goal is to measure the behavior
of the input DNN, so as to provide the input values of the model
needed to run Algorithm~\ref{alg:optBP}, {\it i.e.} the memory sizes
$\wx[\ell], \wbx[\ell], \wy[\ell]$, the memory overheads $\of[\ell],
\ob[\ell]$, and the execution times of each operation in the sequence
$\uf[\ell], \ub[\ell]$.

Parameter estimation is done in the following way: given a chain and a
sample input data $\tilde{a}_0$, forward and backward operations of
each stage are processed one after the other. From $\tilde{a}_{\ell}$
the forward operation $\eforop[\ell]$ is processed to obtain
$\bar{\tilde{a}}_{\ell+1}$, and the backward operation with an
arbitrary value $\delta_{\ell+1}$. The execution time of each
operation is measured, and the memory management interface of PyTorch
is used to obtain the memory usage of $\bar{\tilde{a}}_{\ell+1}$ and
the peak memory usage of both forward and backward operations.

This parameter estimation assumes that the computations performed by
the neural network do not depend on the input data (a very similar
assumption is made for the \texttt{jit.trace()} function of PyTorch),
so that the measurement on a sample input $\tilde{x}$ is
representative of the actual execution on the training data $x$. Adapting
the approach presented in this paper on a data-dependent network would
require both to be able to correctly predict the execution times for
each given input and to recompute the optimal sequence for each new
input, and is thus out of the scope of this paper.

\subsection{Computing the optimal sequence}
Once all measurements have been performed, for any given memory limit
$M$, the optimal persistent sequence can be computed and stored for the
processing phase. In order to limit the computational cost of this
phase, all measured memory sizes are \emph{discretized}: we fix a
number $S$ of memory slots (500 is a reasonable value that we used for
all experiments in this paper), each with size $\frac{M}{S}$, and all
memory sizes are expressed as an integer number of slots, rounded up
if necessary. The complexity of the resulting algorithm is thus
independent of the actual memory limit, at the cost of at most
$1+\frac{1}{S}$ overestimation of memory sizes. We provide a C
implementation of the dynamic programming algorithm, whose running
time on most of the networks in our experiments is below 1 second. The
longest execution time was obtained with ResNet 1001
network~\cite{resnet1001}, which results in a chain of length 339, and
an execution time below 20 seconds. Since this computation is
performed once for the whole training phase, such an execution time is
completely acceptable.

\subsection{Experimental setting}

All experiments presented in this paper are performed with Python
3.7.3 and PyTorch 1.1.0.  The computing node contains 40 Intel Xeon
Gold 6148 cores at 2.4GHz, with a Nvidia Tesla V100-PCIE GPU card with
15.75GB of memory. We experiment with three different kinds on
networks, whose implementation is available in the {\tt torchvision}
package of PyTorch: ResNet, DenseNet, and Inception v3. All three
types of networks have been slightly adapted to be able to use our
tool, by using a \texttt{nn.Sequential} module where applicable. We
use all available depths for ResNet: 18, 34, 50, 101, 152 are
available in {\tt torchvision}, and we also use versions with depth
200 and 1001 proposed in previous work~\cite{resnet1001}. Similarly,
for DenseNet, we use depths 121, 161, 169 and 201.

We use three different image sizes: small images of shape
$224\times224$ (which is the default and minimal image size for all
models of {\tt torchvision}), medium images of shape $500\times500$, and
large images of shape $1000\times1000$. For each model and image size,
we consider different batch sizes that are powers of 2, starting from the smallest
batch size that ensures a reasonable throughput\footnote{With
  small batch sizes, we observe that doubling the batch size
  effectively doubles the throughput, which shows that the GPU is not
  used efficiently in the former case.}.

We compare four strategies to perform a training iteration on those
models:
\begin{itemize}
\item The \textbf{PyTorch} strategy consists in the standard way of
  computing the forward and backward operations, where all
  intermediate activations are stored.
\item The \textbf{sequential} strategy relies on the
  \texttt{checkpoint\_sequential} tool of
  PyTorch~\cite{periodiccheckpointing}. This strategy splits the chain
  into a given number of segments $s$ and, during the forward phase
  only, stores activations at the beginning of each segment. Each
  forward computation is thus performed twice, except those of the
  last segment. We use $10$ different number of segments, from $2$
  (always included) to $2\sqrt{L}$, where $L$ is the length of the
  chain\footnote{Note that $\sqrt{L}$ is the optimal number of
    segments for this strategy when the chain is homogeneous.}. The
  same strategy is used in~\cite{memonger}, but the number of segments
  needs to be hand-tuned.
\item The \textbf{revolve} strategy uses the optimal algorithm adapted
  to heterogeneous chains of the Automatic Differentiation
  model~\cite{griewank2008evaluating}, and converts it to a valid
  solution by saving only activations $a$ to memory, and performing a
  $\eforop$ step before each backward step to enforce validity. This
  is the same strategy as advocated in Appendix C
  of~\cite{gruslys2016memory}.
\item The \textbf{optimal} strategy uses Algorithm~\ref{alg:optBP} for
  $10$ different memory limits, equally spaced between $0$ and the
  memory usage of the \textbf{PyTorch} strategy.
\end{itemize}

For each model, image size and batch size, we perform enough
iterations to ensure that the \textbf{PyTorch} strategy lasts at least
$500$ms, and we measure the actual peak memory usage and duration over
$5$ runs. The obtained measurements are very stable, so all plots in
the next section present the median duration over the $5$ runs for
each experiment (on average, the difference between the highest and
lowest measured throughput is 0.5\% of the median).  For each run, the
memory peak consumption and the throughput of the experiments have
been carefully assessed, using the same mechanism as the one used to
perform the measurement phase. The measured values are very close to
the predictions from our model: over all experiments, the mean
absolute percentage error is 7.8\% for throughput, and 3.7\% for peak
memory consumption.

\begin{figure}
  \begin{center}
    \includegraphics[width=\linewidth]{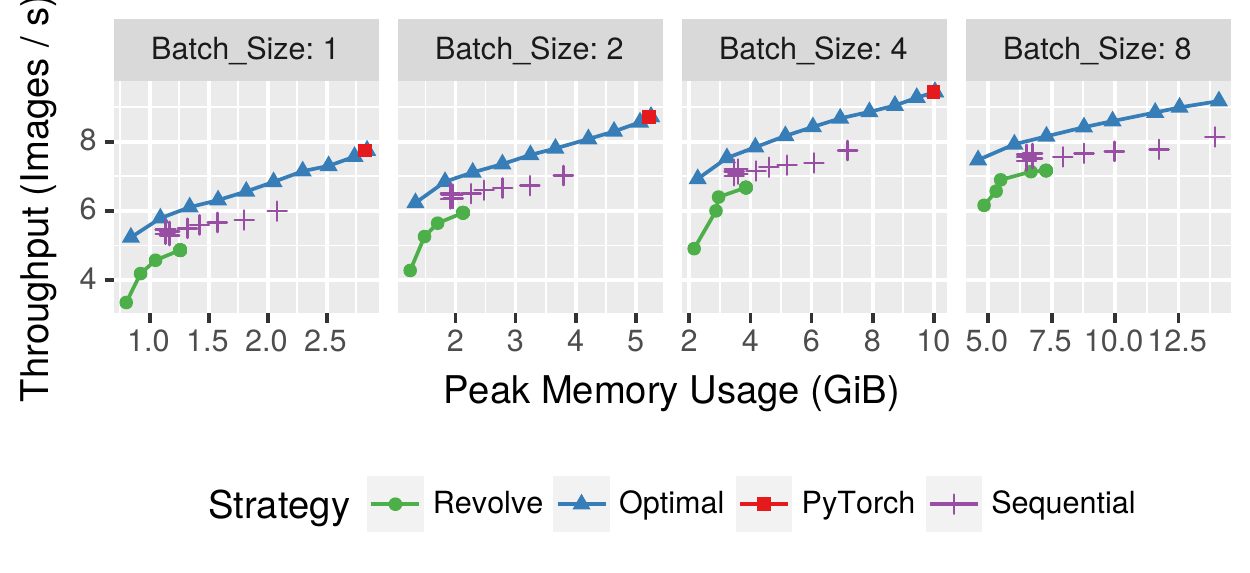}
  \end{center}
  \caption{Experimental results for the ResNet network with depth 101
    and image size 1000.}
  \label{plot:ResNet101}
\end{figure}

\subsection{Experimental Results}
\begin{figure}
  \begin{center}
    \includegraphics[width=\linewidth]{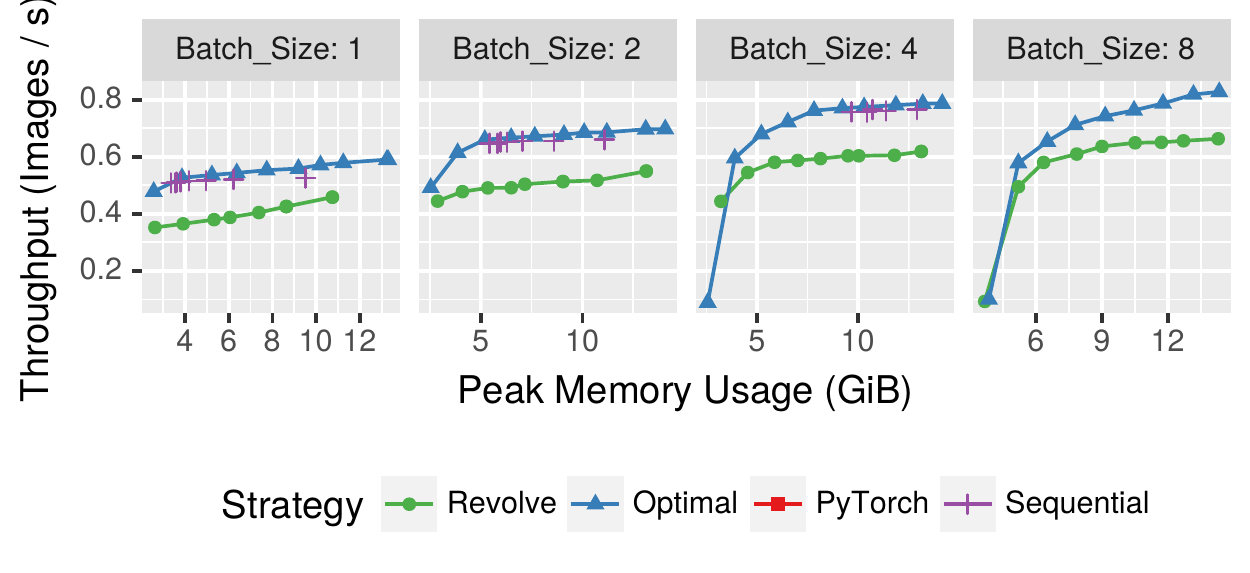}
  \end{center}
  \caption{Experimental results for the ResNet network with depth 1001
    and image size 224.}
  \label{plot:ResNet1k}
\end{figure}

\begin{figure}
  \begin{center}
    \includegraphics[width=\linewidth]{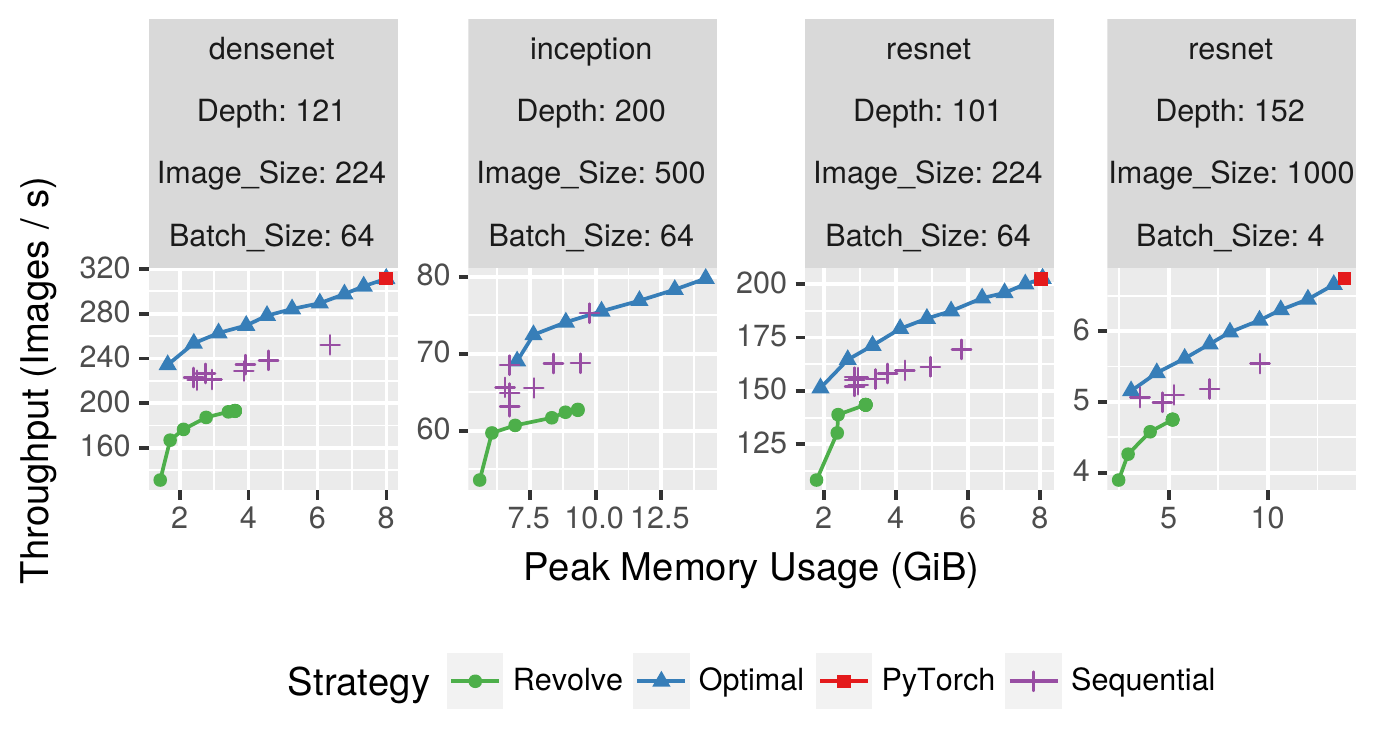}
  \end{center}
  \caption{Experimental results for several situations.}
  \label{plot:mix}
\end{figure}

\begin{figure}[p]
  \begin{center}
    \includegraphics[width=\linewidth]{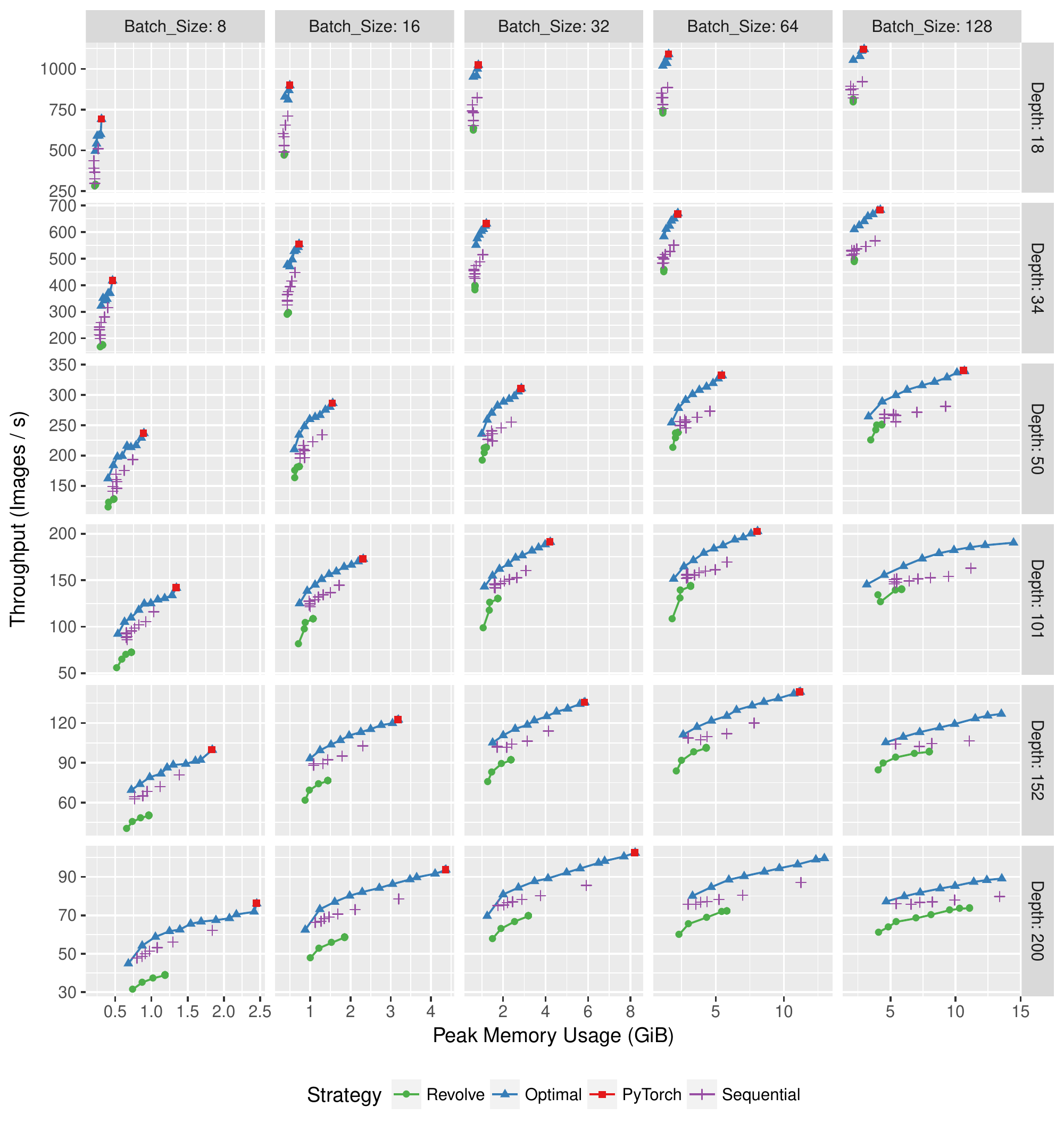}
  \end{center}
  \caption{Results for Resnet with image size 224, for different
    depths and batch sizes.}
  \label{plot:resnet:224}
\end{figure}

\begin{figure}[p]
  \begin{center}
    \includegraphics[width=\linewidth]{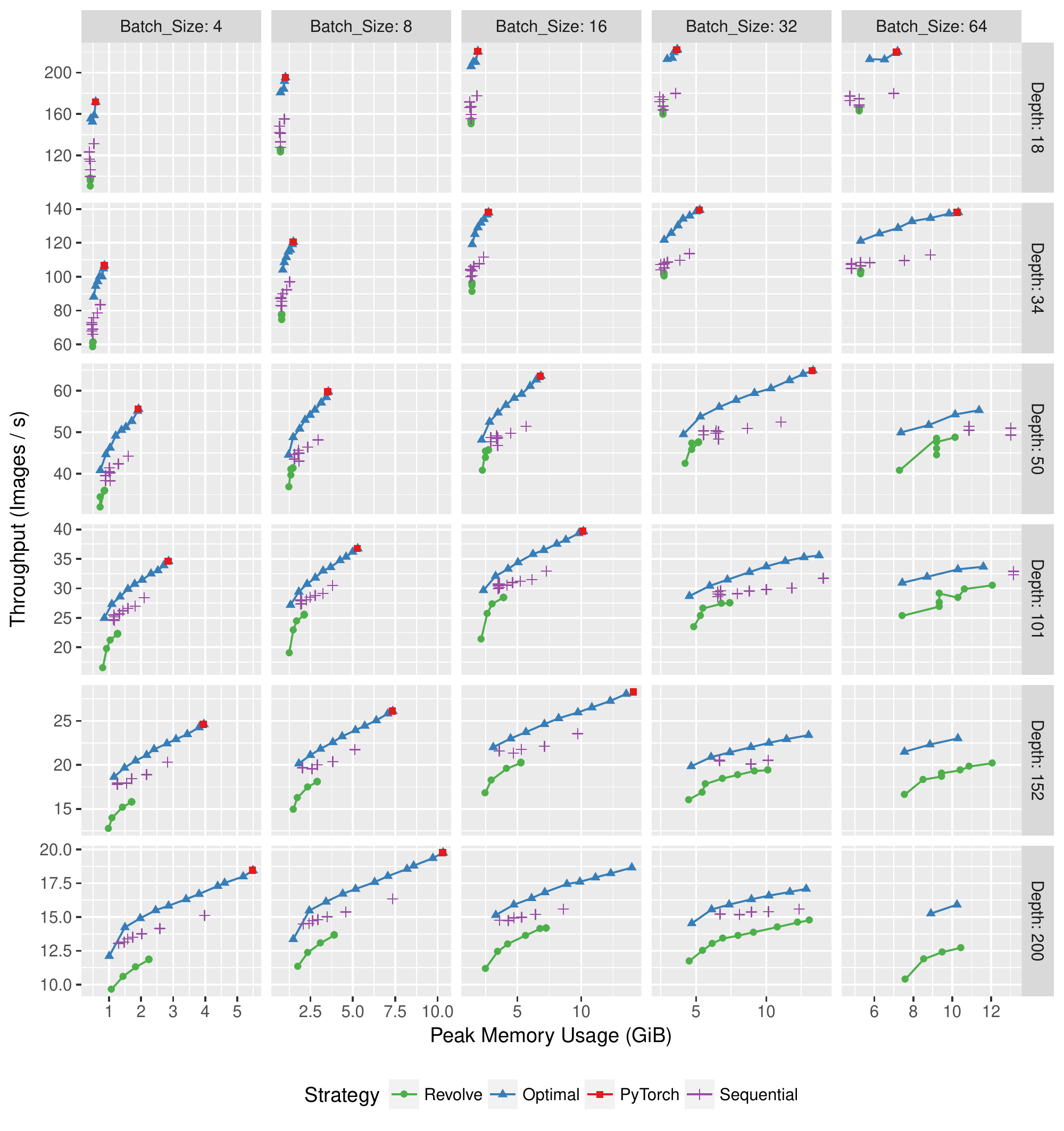}
  \end{center}
  \caption{Results for Resnet with image size 500, for different
    depths and batch sizes.}
  \label{plot:resnet:500}
\end{figure}

\begin{figure}[p]
  \begin{center}
    \includegraphics[width=\linewidth]{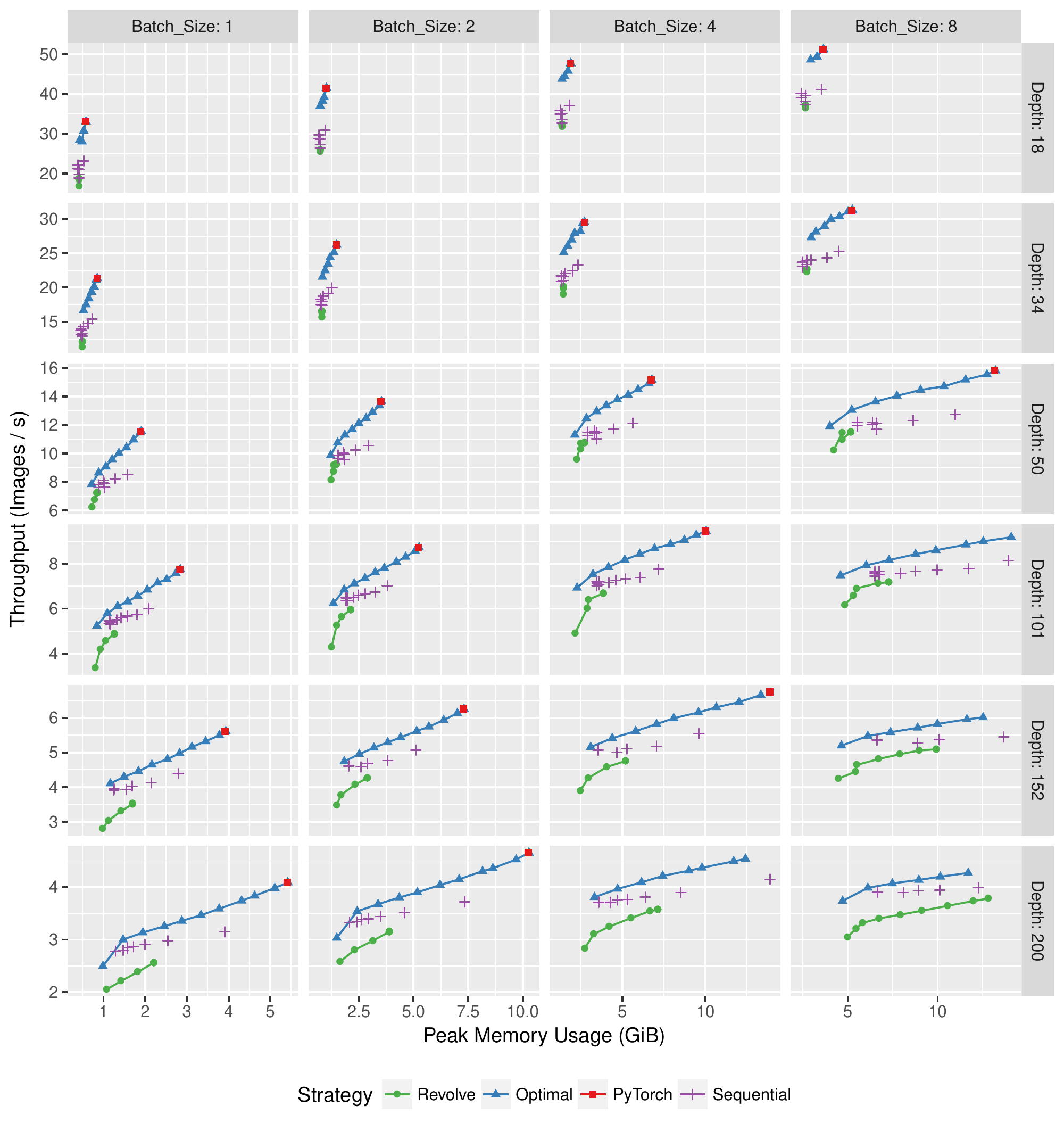}
  \end{center}
  \caption{Results for Resnet with image size 1000, for different
    depths and batch sizes.}
  \label{plot:resnet:1000}
\end{figure}

\begin{figure}[p]
  \begin{center}
    \includegraphics[width=\linewidth]{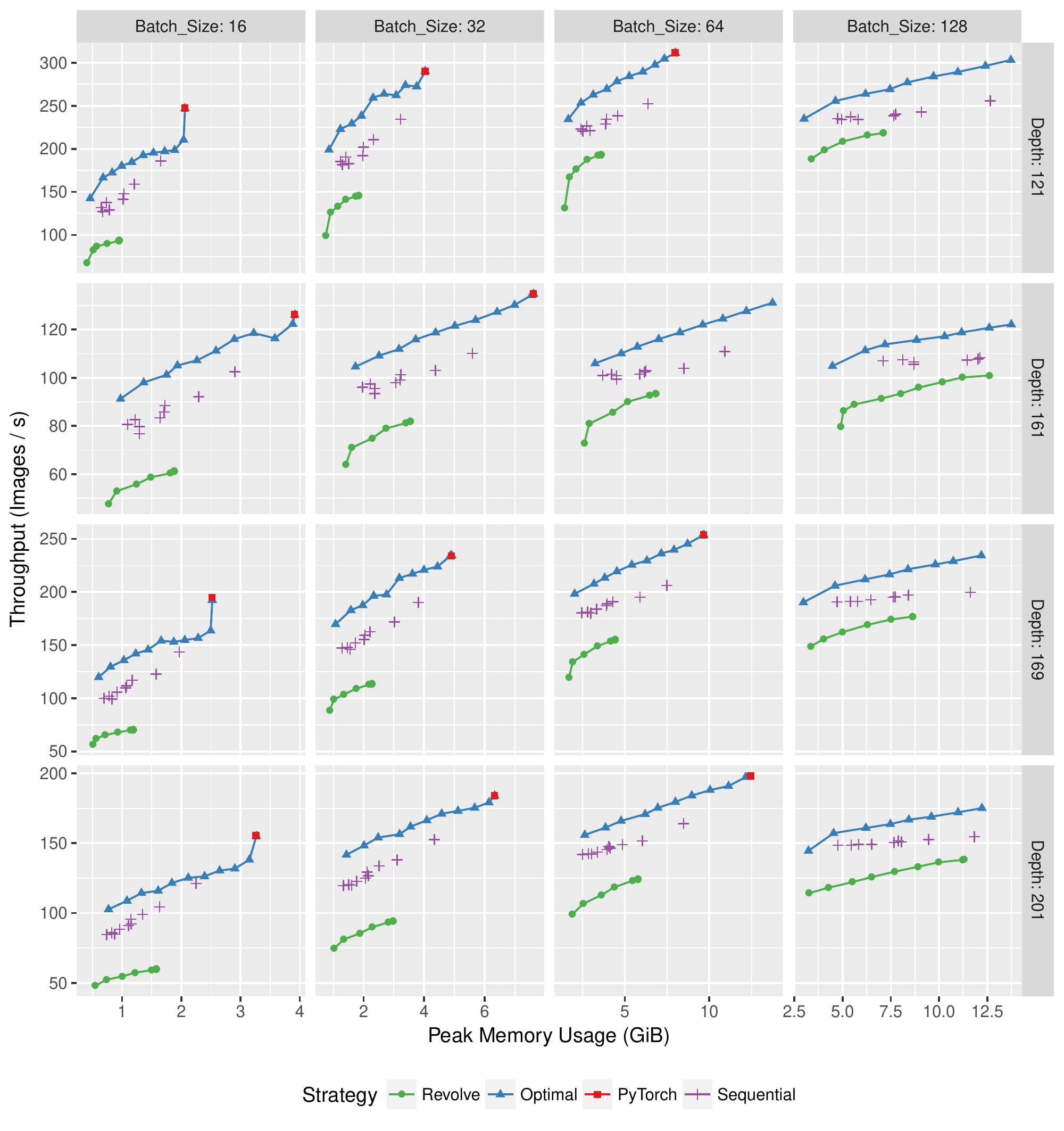}
  \end{center}
  \caption{Results for Densenet with image size 224, for different
    depths and batch sizes.}
  \label{plot:densenet:224}
\end{figure}

\begin{figure}[p]
  \begin{center}
    \includegraphics[width=\linewidth]{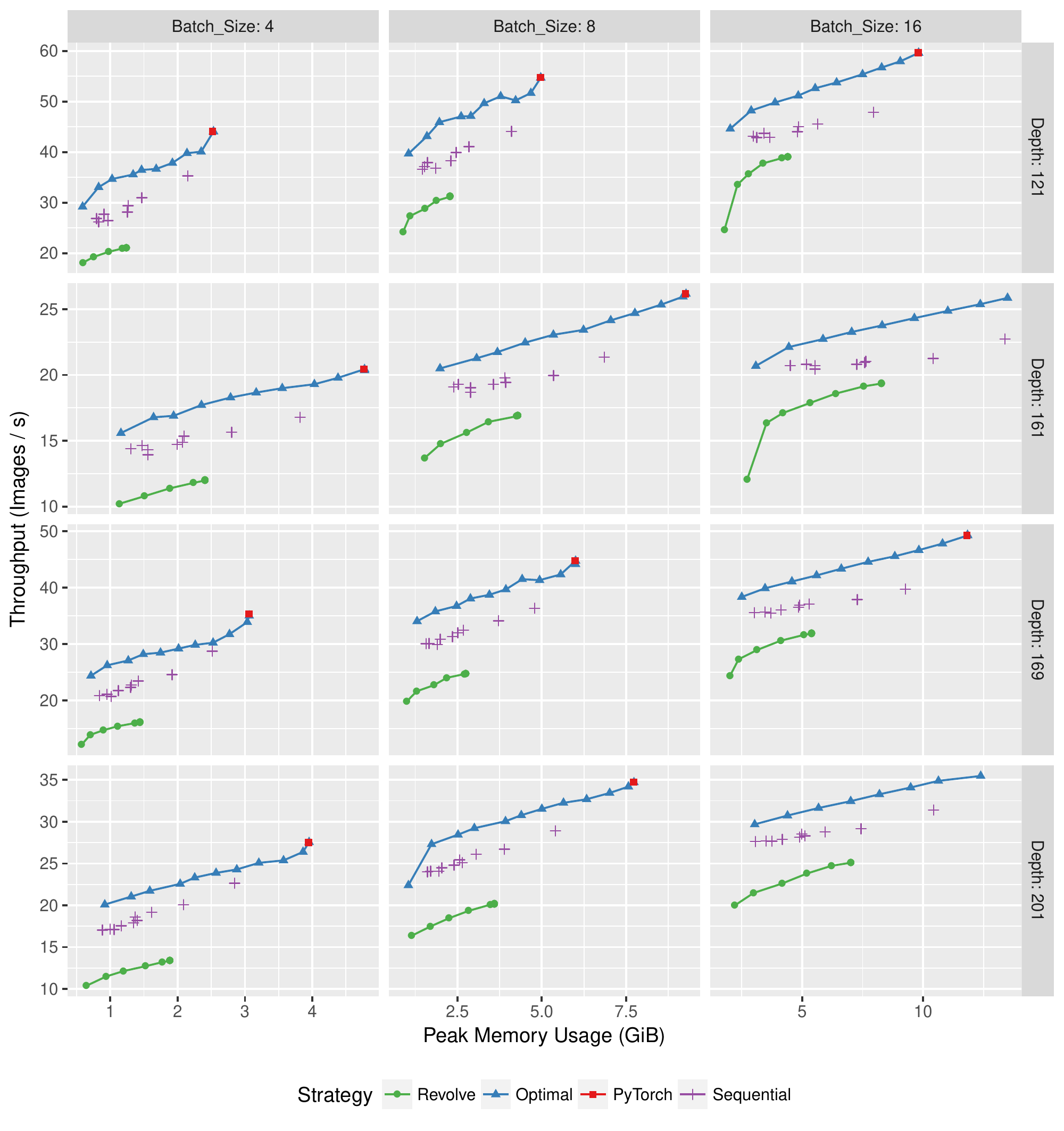}
  \end{center}
  \caption{Results for Densenet with image size 500, for different
    depths and batch sizes.}
  \label{plot:densenet:500}
\end{figure}

\begin{figure}[p]
  \begin{center}
    \includegraphics[width=\linewidth]{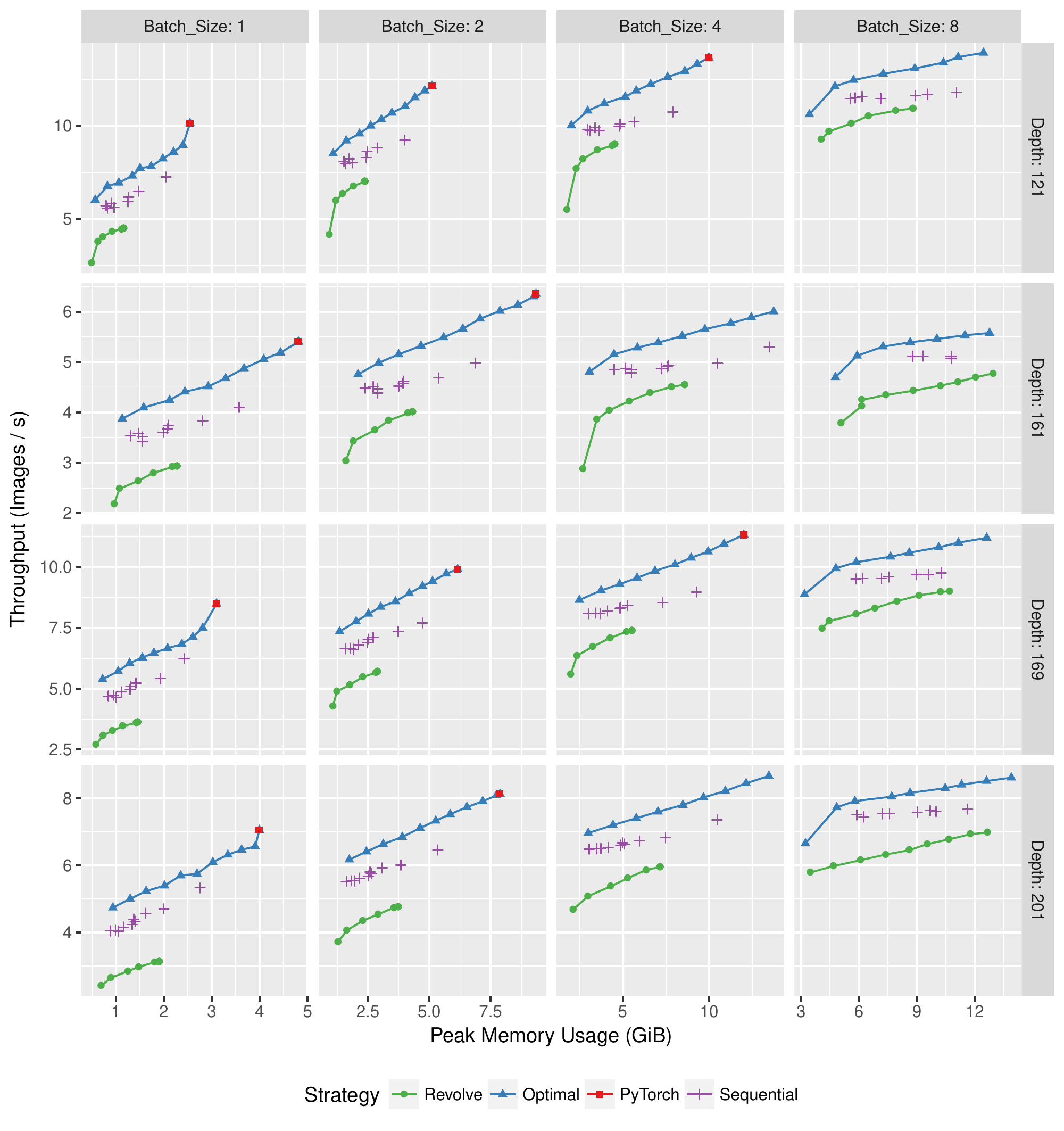}
  \end{center}
  \caption{Results for Densenet with image size 1000, for different
    depths and batch sizes.}
  \label{plot:densenet:1000}
\end{figure}

\begin{figure}[p]
  \begin{center}
    \includegraphics[width=\linewidth]{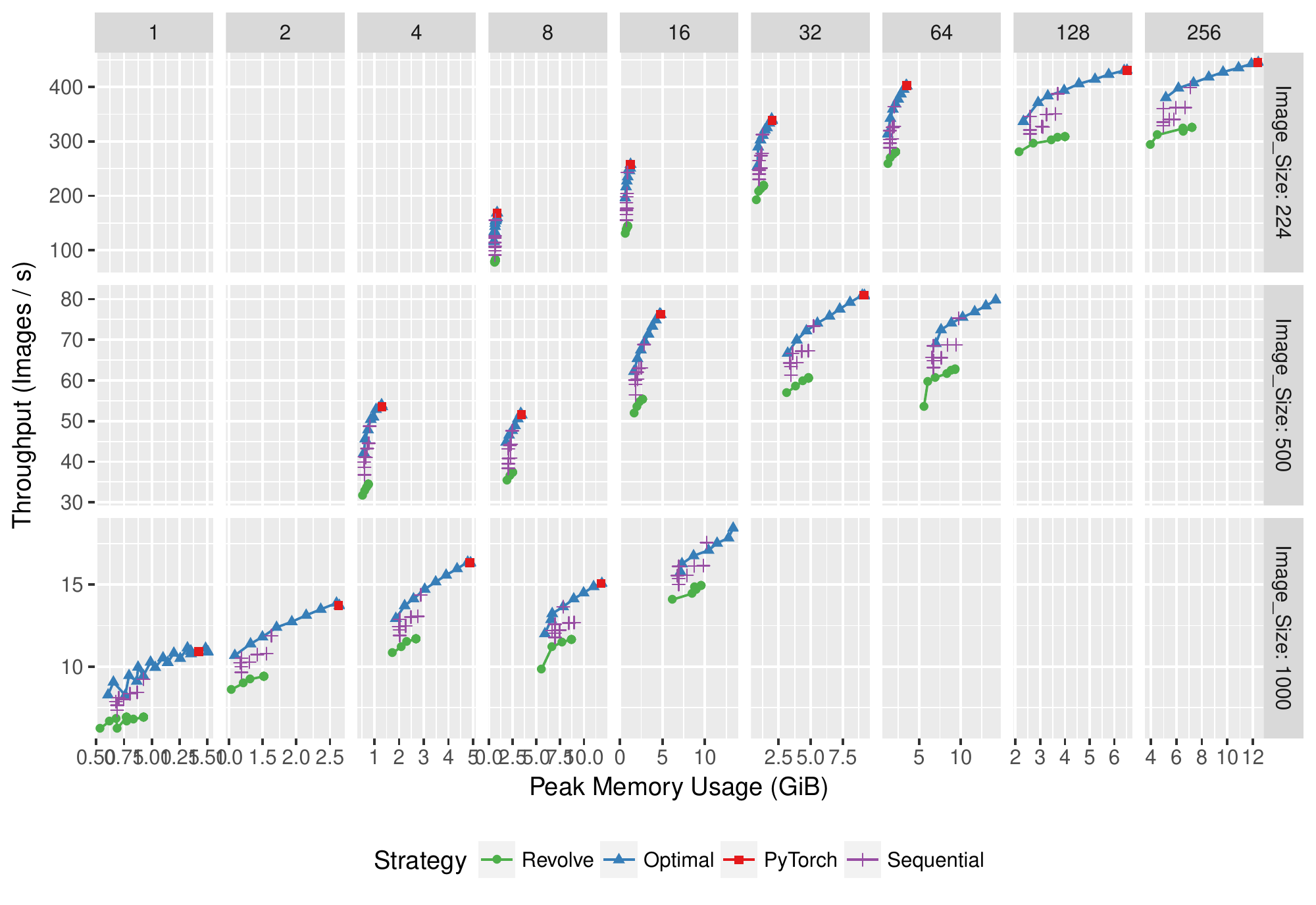}
  \end{center}
  \caption{Results for Inception v3 for different image sizes and
    batch sizes.}
  \label{plot:inception}
\end{figure}

\begin{figure}[p]
  \begin{center}
    \includegraphics[width=\linewidth]{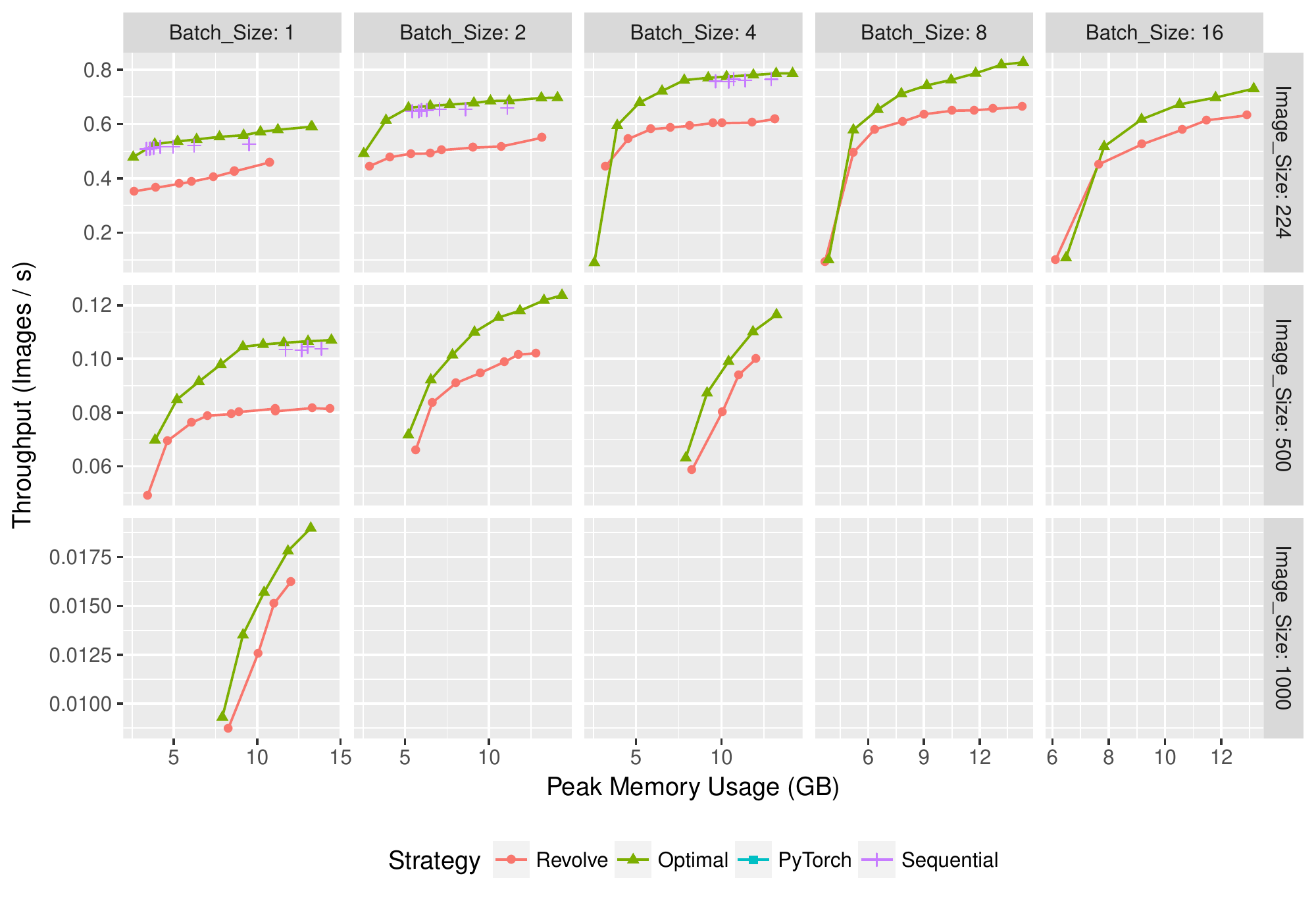}
  \end{center}
  \caption{Results for Resnet 1001, for different image sizes and
    batch sizes.}
  \label{plot:resnet1k}
\end{figure}

All plots corresponding to above described experiments are available
in the supplementary material. For the sake of conciseness, we only
present here a representative selection of the results; the behavior
on other experiments is very similar. All plots have the same
structure: for a given set of parameters (network, depth, image size
and batch size), we plot for each strategy the achieved throughput (in
terms of images per second) against the peak memory usage.
%% This was in supplementary, remove if space is needed.
The square red dot represents the performance obtained by the
standard \textbf{PyTorch} strategy, and its absence from the graph
means that a memory overflow error was encountered when attempting to
execute it. Purple crosses represent the results obtained with the
\textbf{sequential} strategy for different number of segments. The
blue line with triangles shows the result obtained with our
\textbf{optimal} strategy. The green line with circles show the result
obtained with the \textbf{revolve} algorithm. We draw lines to
emphasize the fact that these strategies can be given any memory limit
as input, whereas the result of \textbf{sequential} is inherently tied
to a discrete number of segments. We provide a representative
selection of results in Figures~\ref{plot:ResNet101}
to~\ref{plot:mix}, and the complete results can be found in
Figures~\ref{plot:resnet:224} to~\ref{plot:resnet1k} at the end of the
paper.

%\begin{figure}
%  \begin{center}
%    \includegraphics[width=0.8\linewidth]{Figure4.pdf}
%  \end{center}
%  \caption{Experimental results for the Resnet network, with image size 1000, for various depths and batch sizes}
%  \label{plot:ResNet:mix}
%\end{figure}

Figure~\ref{plot:ResNet101} shows the results for the ResNet neural
network with depth of $101$, with image size $1000 \times 1000$ and
batch size $1$, $2$, $4$, and $8$. For a batch size of $1$,
\textbf{PyTorch} strategy has a memory peak consumption of $2.83$ GiB
which is enough to fit on this GPU.  However, when the batch size is
$8$, \textbf{PyTorch} strategy fails to compute the back-propagation
due to memory limitations. The \textbf{sequential} strategy offers a
discrete alternative by dividing the chain into a given number of
segments (in this case from $2$ to $11$).  For every batch size, the
best throughput is reached when the number of segments is equal to
$2$. For instance, when the batch size is $8$, the throughput of the
\textbf{sequential} strategy with $2$ segments is on average $8.13$
images/s with a memory peak consumption of $13.91$ GiB. The
\textbf{optimal} strategy offers a continuous alternative by
implementing the best checkpointing strategy for any given memory
bound. We can see that for a given memory peak, the \textbf{optimal}
strategy outperforms the \textbf{sequential} strategy by 15\% on
average. For instance, when the batch size is $8$, the maximum
throughput achieved by the \textbf{optimal} strategy is $9.18$
images/s. The previous \textbf{revolve} algorithm provides a
continuous approach as well. However, it requires to compute each
forward operation at least twice (once in the forward phase, once
before the backward operation), which incurs a much lower throughput
than both other solutions. Furthermore, since this algorithm does not
consider saving the larger $\bar{a}$ values, it is unable to make use
of larger memory sizes.

Figure~\ref{plot:ResNet1k} displays the same results for the ResNet
with depth of $1001$ and image size of $224 \times 224$. This setup
requires much more memory and the \textbf{PyTorch} strategy fails even
when the batch size is $1$. The \textbf{sequential} strategy requires
at least $6$ segments for batch size $1$, $10$ segments for batch size
$2$, and $18$ segments for batch size $4$, and cannot perform the
back-propagation when the batch size is $8$.  Not only does the
\textbf{optimal} strategy outperform the \textbf{sequential} strategy
when it does not fail but it offers a stable solution to train the
neural network even with a larger batch size, which allows to increase
the achieved throughput thanks to a better GPU efficiency (0.82 for
\textbf{PyTorch} whereas the highest throughput achieved by
\textbf{sequential} is 0.76).  It is interesting to note that based on
the parameters estimated by our tool, running the setting with batch
size 8 with the \textbf{PyTorch} strategy would require 225 GiB of
memory, and achieve a throughput of 1.1 images/s. Additional results
in Figure~\ref{plot:resnet1k} also show that \textbf{optimal} allows
to run this large network even with medium and large image sizes.

All these conclusions hold for every tested neural network and
parameters. Figure~\ref{plot:mix} displays some of them and shows that the
behavior of the \textbf{optimal} strategy is stable on various network
sizes and image sizes. To summarize, we also compute the ratio between
the highest throughput obtained by \textbf{sequential} and the
throughput achieved by \textbf{optimal} with the corresponding memory
usage. On average over all tested sets of parameters, \textbf{optimal}
achieves 17.2\% higher throughput.
%% result of summarize.py in code/plots/AAAI

\section{Conclusion}
\label{sec.conclusion}

This document describes a new checkpointing strategy that leverages
operations available in DNN frameworks with the capabilities of
autograd functions. We carefully model back-propagation and we propose
a dynamic programming algorithm which computes the optimal persistent
schedule for any sequentialized network and its implementation for any
sequential Pytorch module. Using in-depth experiments, we compare
achieved results against (i) a periodic checkpointing strategy
available in PyTorch and (ii) an optimal persistent strategy adapted
from the Automatic Differentiation literature to a fully heterogeneous
setting, but which does not use all the capabilities available in DNN
frameworks. We show that our implementation consistently outperforms
these two checkpointing strategies, for a large class of networks,
image sizes and batch sizes. Our fully automatic tool increases
throughput by an average of 17.2\% compared to its best competitor,
with better flexibility since it offers the ability to specify an
arbitrary memory limit.  Our tool therefore allows you to use larger
models, larger batches or larger images while automatically adapting
to the memory of the training device.  In our future work, we want to
study the advantages of our approach in combination with other
strategies developed to address memory limitations such as model
parallelism, activation offloading and domain decomposition.

\bibliography{mybib}
\bibliographystyle{acm}

\end{document}